\documentclass[sigconf]{acmart}

\usepackage{graphicx}
\usepackage{stfloats}
\usepackage{amsmath}
\usepackage{multirow}
\usepackage[ruled,linesnumbered]{algorithm2e}
\usepackage{bm}
\usepackage{booktabs}
\usepackage[switch]{lineno}
\usepackage{tabularx}
\usepackage{microtype}

\usepackage{algorithm}
\usepackage{subfigure}
\usepackage{multirow}
\usepackage{hhline}
\usepackage{xcolor}
\usepackage{enumitem}

\usepackage{makecell}

\AtBeginDocument{%
  }

\renewcommand\footnotetextcopyrightpermission[1]{} 
\begin{document}



\title{DySENSE: Dynamic Spatiotemporal ENergy Synthesis and Emulation with Satellite Imagery}



\author{Kailai Sun}
\orcid{0000-0003-1648-3409}
\affiliation{%
  \institution{SMART, Singapore}
  \country{ }
}
\affiliation{%
  \institution{Massachusetts Institute of Technology}
  \city{Cambridge}
  \state{MA}
  \country{USA}
}
\email{skl24@mit.edu}

\author{Zijie Gao}
\affiliation{%
  \institution{National University of Singapore}
  \country{Singapore }
}
\email{E1349501@u.nus.edu}

\author{Mingyi He}
\affiliation{%
  \institution{Massachusetts Institute of Technology}
  \city{Cambridge}
  \state{MA}
  \country{USA}
}
\email{mingyihe@mit.edu}

\author{Can Rong}
\affiliation{%
  \institution{SMART, Singapore}
  \country{}
}
\affiliation{%
  \institution{Massachusetts Institute of Technology}
  \city{Cambridge}
  \state{MA}
  \country{USA}
}
\email{rongcan@mit.edu}

\author{Alok Prakash}

\affiliation{%
  \institution{SMART, Singapore}
  \country{}
}
\affiliation{%
  \institution{Massachusetts Institute of Technology}
  \city{Cambridge}
  \state{MA}
  \country{USA}
}
\email{alokprks@mit.edu}

\author{Baoshen Guo}
\affiliation{%
  \institution{SMART, Singapore}
  \country{}
}
\affiliation{%
  \institution{Massachusetts Institute of Technology}
  \city{Cambridge}
  \state{MA}
  \country{USA}
}
\email{baoshen@mit.edu}

\author{Shenhao Wang}
\authornote{Corresponding author.}
\orcid{0000-0003-4374-8193}
\affiliation{%
  \institution{University of Florida}
  \city{Gainesville}
  \state{FL}
  \country{USA}
}
\email{shenhaowang@ufl.edu}

\author{Jinhua Zhao}
\authornotemark[1]
\orcid{0000-0002-1929-7583}
\affiliation{%
  \institution{Massachusetts Institute of Technology}
  \city{Cambridge}
  \state{MA}
  \country{USA}
}
\email{jinhua@mit.edu}

\renewcommand{\shortauthors}{Kailai et al.}

\begin{abstract}
Cities account for over two-thirds of the world’s energy and more than 70\% of global carbon dioxide emissions. Urban Building Energy Modelling (UBEM) is essential for achieving the 2050 net-zero target and the UN's Sustainable Development Goals 7 and 11. Although existing studies have made progress, they are limited by data and methodological gaps. Existing UBEM is largely predictive for existing cities rather than generative for future cities; fine-grained, aligned monthly dynamic energy data are scarce and limited; and existing GenAI models struggle to generate dynamic urban functional monthly energy. We present DySENSE, a joint multimodal generative AI framework that synthesizes static urban satellite imagery together with spatially aligned, dynamic monthly building energy consumption maps. We decouple long-term static urban morphology from short-term dynamic energy consumption. By conditioning on static urban land use, DySENSE can generate realistic, diverse and new static urban imagery with aligned energy maps. By conditioning on dynamic monthly and climate prompts, DySENSE can jointly generate dynamic, monthly energy consumption maps while keeping the urban morphology static. Experiments across four cities (New York City, Washington, D.C., Norfolk, and Busan) demonstrated that DySENSE can generate annotated high-fidelity synthetic energy data using less than 25\% labelled data, boosting downstream prediction performance by 8\% mIoU, and achieving strong energy dynamic consistency. Compared to existing energy prediction methods, DySENSE significantly improved prediction performance (about 8.7\% accuracy) and reduced prediction error (9\%-18\% CVRMSE, ASHRAE metrics). DySENSE offers an effective foundation to support energy-efficient urban planning, peak-load analysis, building performance simulation and dynamic demand response for urban, energy, and building science. The dataset, model and code: \url{https://huggingface.co/datasets/skl24/MUSE2} and \url{https://github.com/kailaisun/DySENSE}.
\end{abstract}

\ccsdesc[500]{Computing methodologies~Artificial intelligence}
\vspace{-1em}

\keywords{Urban Building Energy Modeling; Diffusion Models; Spatiotemporal Dynamics;
Earth Observation; Data Augmentation}


\maketitle

\section{Introduction}

According to United Nations (UN) data, 55\% of the global population lives in cities, a share projected to reach 68\% by 2050 \cite{UN2018Urbanization}. With the rapid urbanisation of the global population, cities have become the key contributors to energy and climate challenges. According to the IEA \cite{IEA2021Cities}, cities account for more than two-thirds of global energy consumption and more than 70\% of global carbon emissions. To achieve net-zero emissions goals by 2050 \cite{IEA2025WEO}, Urban Building Energy Modelling (UBEM) plays \cite{wang2023impacts,johari2020urban} an essential role, as buildings consume about 32\% of global energy and contribute to 34\% of global CO2 emissions \cite{GlobalABC2024}.

UBEM is a fundamental tool for urban energy performance analysis, enabling effective management and optimisation of energy systems \cite{zhou2025state,li2025integrating}. Effective modelling and planning of urban energy dynamics are important for policy-making and peak load assessments, advancing the UN Sustainable Development Goals (SDGs) 7 (Affordable and Clean Energy) and 11 (Sustainable Cities and Communities)  \cite{zhou2025state,sun2026vision,yap2025revealing}. UBEM approaches can be divided into top-down and bottom-up approaches \cite{li2025integrating,ali2021review,sezer2023urban}. Top-down models provide a macro-scale perspective for assessing or predicting building energy \cite{ghorbany2025systematic}, connecting aggregate urban energy consumption to socioeconomic, statistical, and remote-sensing variables \cite{wang2023impacts}, such as population, GDP, land use, temperature, and nighttime-light data \cite{li2026satellite}. In contrast, bottom-up models describe individual buildings or building clusters and then aggregate their energy performance at the city scale \cite{chen2025building,hong2020ten}, relying on building morphology, materials, Heating, Ventilating and Air-Conditioning (HVAC) systems, occupancy schedules, weather, and building functions \cite{ali2021review,deng2023autobps}. 

Recently, UBEM studies have utilized satellite imagery as an important tool for urban monitoring and energy prediction, as satellite imagery provides rich spatial and temporal information \cite{xiao2024refound,sun2025flexireg}.  \citet{li2026xai} combined multi-scale satellite imagery with tabular building and weather data and CNN models to estimate monthly site energy use intensity in  Washington, D.C., while \citet{yap2025revealing} integrated multimodal geospatial inputs and deep learning to quantify building operating carbon emissions across five cities. \citet{MAYER2023120542}  applied aerial imagery, street-view imagery, and temperature data to estimate building energy efficiency using Inception-3 models in the UK. \citet{sun2026deciphering} fused street-view imagery, aerial imagery, thermal infrared imagery, building morphological attributes, and socioeconomic indicators to estimate building energy efficiency using Swin Transformer-based models in Glasgow and Edinburgh. \citet{WANG2025106054} proposed sat2shp, which uses Mask R-CNN to infer 2.5D building massing and use type from satellite imagery for UBEM in Chicago and San Francisco. \citet{streltsov2020estimating} trained CNNs to segment residential and commercial buildings and then predict energy consumption using overhead imagery.  Night-time light imagery has further been used to infer spatial patterns of building energy use and carbon emissions \citep{FEHRER2018252,wang2024estimation,li2026satellite}.

With the rapid development of generative AI and remote sensing \cite{10179972,zhao2024artificial}, recent studies have begun to explore diffusion models \cite{ho2020denoising} in urban \cite{chai2024diffusion} and energy domains. DiffusionSat \cite{khanna2024diffusionsat} proposed a GenAI foundation model from Stable Diffusion (SD) and latent diffusion models (LDMs) \cite{rombach2022high} for satellite imagery generation using remote sensing metadata. \citet{zhou2026geometric} develop a diffusion model to generate white and black maps of city buildings conditioned on text descriptions.  \citet{WANG2025102339} and \citet{HE2026113892} generated high-fidelity satellite imagery and urban layouts for urban planning scenarios in Chicago, Dallas, and Los Angeles. For the UBEM domain, \citet{sun2026sense} linked generative satellite imagery to building height and static annual energy consumption. \citet{fu2024creating} use a conditional diffusion model to generate 1D building energy meter data from building metadata and weather. 

Although existing UBEM studies have made progress, they are limited by data and methodological gaps. First, most UBEM approaches are primarily predictive and deterministic, estimating energy consumption based on existing urban plans, building characteristics, and environmental conditions. They struggle to generate or envision new, energy-efficient urban morphologies for future cities. Second,  a key data challenge in the data-driven UBEM is the scarcity of aligned, high-quality and dynamic monthly building energy data due to privacy, cost, sensitivity, etc \cite{zhou2025state}. But the dynamic monthly energy data is important for operational assessments, peak load analysis, building performance simulation, and energy demand response \cite{wang2023impacts}. Third, although existing generative models can generate the visual modality (RGB) of urban imagery, they lack the corresponding generation of monthly energy dynamics.



To address the above gaps, we propose DySENSE, a joint multi-modal generative AI framework for both static urban satellite imagery and dynamic building energy consumption generation. Unlike static annual energy generation \cite{sun2026sense}, DySENSE is a \textbf{dynamic} satellite-based monthly and climate-aware energy synthesis framework. In urban science, urban morphology evolves over long horizons and embodies the cumulative effects of long-term urban development \cite{zhang2026building}, but building energy consumption significantly differs across months and climates \cite{wang2023impacts}. Based on this, we decouple the generative process into static and dynamic branches. By conditioning on static textual urban land-use metrics, DySENSE can generate realistic, diverse and new static urban imagery, with aligned high-resolution building energy consumption maps. By conditioning on dynamic monthly and climate prompts, our framework can jointly generate dynamic, monthly energy consumption maps for a full year, while keeping the same static urban morphology. On the other hand, we collect monthly fine-grained building energy data across multiple cities worldwide and perform spatiotemporal alignment with climate, satellite imagery, and land use data. Our framework is validated on this dataset, which covers Busan, New York City, Norfolk, and Washington, D.C. The main contributions:
\begin{itemize}
    \item We propose a joint generative AI framework for generating both static urban satellite imagery and dynamic monthly, aligned building energy maps, conditioned on static textual urban land-use information and dynamic monthly and climate prompts.
    \item We establish a multi-city satellite and monthly energy consumption dataset covering NYC, Busan, Norfolk, and Washington, D.C., where monthly energy disclosure records are spatiotemporally aligned with satellite imagery, climate data, and land-use information.
    \item By learning the joint distribution, our framework can jointly generate both RGB and energy modalities using the RGB VAE and the proposed Energy VAE; by inferring the conditional distribution, it can generate aligned dynamic building energy maps from generated static satellite imagery. 
    \item We conduct many evaluation experiments, which demonstrate that our generative data augmentation strategy improves energy prediction performance by about 8\% mIoU using less than 25\% of the real training data.
    \item Compared to existing multi-modal urban building energy prediction methods, our method significantly reduced energy prediction error (reduced 9\%-18\% CVRMSE).
\end{itemize}
\section{Monthly Multi-city Urban Satellite-Energy Dataset}




  \subsection{Data Coverage}

  We establish a monthly multi-city dataset, as defined by the GHS Urban Centre Database \cite{ghs_ucdb}, covering four cities across North America and East Asia: New York City, Washington, D.C., Norfolk, and Busan. For each city, we align
  monthly municipal-scale building energy disclosure records with satellite imagery, land-use text prompts at a fixed spatial extent, in Tab. \ref{tab:city_dataset_stats}. Each sample corresponds to a $2\,\mathrm{km}\times 2\,\mathrm{km}$ tile, represented by: (1) a satellite image, (2) a text prompt describing urban land use and month, (3) a building-level energy map where the energy values are transformed by a log1p function.

  \subsection{Dataset Overview}

  \subsubsection{Satellite Imagery and Land Use}

  Urban boundary data are obtained from the Global Human Settlement (GHS) Urban Centre Database 2023~\cite{ghs_ucdb}.    High-resolution satellite imagery was obtained from Mapbox basemaps \cite{mapbox_satellite_imagery} and cropped into $512 \times 512$ pixel images for each $2\,\mathrm{km}\times 2\,
  \mathrm{km}$ tile. Urban morphology and land-use attributes are derived from geospatial layers associated with each city. For each tile, we construct a natural-language prompt
  describing the city, month, and land-use composition, such as residential, commercial, industrial, service, parking, park, and natural land use.
  

\vspace{-0.5em}
  \subsubsection{Building Energy Consumption}

We compiled monthly building energy data from city-specific public or administrative sources in 2023. For NYC, we leveraged Local Law 84 building energy disclosure records~\cite{nyc_ll84}.  For Norfolk, we leveraged municipal address-level monthly energy records from Norfolk opendata  ~\cite{Norfolk_energy}.   For Busan, we leveraged Korean administrative monthly building energy data ~\cite{busan_city} from the Korean architecture HUB.  For Washington, D.C., we leveraged municipal building energy benchmarking records ~\cite{DC_energy}.

  For each city, monthly electricity and natural gas consumption are converted to a common energy unit, kBtu, and combined as total operating energy for each 100 m $\times$ 100 m cell. Because the energy data (kBtu) exhibit a long-tailed distribution, we apply the log1p function to transform the data into a Gaussian-like distribution. The energy label is formulated as a multi-class map, where class 0 denotes background or no positive energy value, and the remaining 9 classes represent positive energy levels partitioned into equal-frequency bins by quantile thresholds.

  \subsubsection{Climate Data}For each 2 km $\times$ 2 km tile and month, we use five climate variables in the dynamic prompt: monthly mean daily maximum air temperature ($T_{\max}$, $^\circ$C), monthly mean daily minimum air temperature ($T_{\min}$, $^\circ$C), monthly mean daily precipitation (mm), monthly mean daily water vapor pressure (Pa), and monthly mean daily surface shortwave radiation energy (kWh m$^{-2}$).
  
  For NYC, Washington, D.C., and Norfolk, climate variables were derived from Daymet V4, which provides daily gridded weather information at a 1 km spatial resolution \cite{daymet_v4}.  Daily $T_{\max}$, $T_{\min}$, precipitation, water vapor pressure, and day length data were first temporally aggregated to monthly means of daily values.  Surface shortwave radiation was converted from shortwave flux (W m$^{-2}$) to daily radiation energy (kWh m$^{-2}$) using the day length before monthly averaging. Then they are spatially mapped to our 2 km × 2 km tiles using area-weighted averaging. 
  
  For Busan, we used climate data from TerraClimate and the Korea Meteorological Administration (KMA). Monthly $T_{\max}$, $T_{\min}$, and vapor pressure (converted to Pa) were obtained from TerraClimate (about 4.6 km resolution) \cite{terraclimate} via area-weighted averaging of pixels overlapping our 2 km × 2 km tiles. Precipitation (500 m) and surface shortwave radiation (GK2A Level-2 SWRAD, 2 km) were obtained from KMA \cite{kma_api_hub}, spatially area-weighted to each 2 km × 2 km tile and temporally aggregated to monthly mean daily values.


  \subsection{Data Pre-processing}

  We perform spatial and temporal alignment across satellite imagery, text prompts, and building energy records using a unified geospatial coordinate system. Each sample is indexed by city, month, and $2\,\mathrm{km}\times 2\,\mathrm{km}$ tile identifier, ensuring that the satellite image and energy label are spatially and temporally aligned.

We recruited three urban domain specialists to manually review the energy annotations, flagging a tile as unaccepted if its energy label map exhibited large contiguous blocks of missing or null values. We use expert-in-the-loop filtration to ensure that the model is trained on high-quality samples where the spatial distribution of the energy label map aligns with the observed urban morphology. Finally, the high-quality dataset comprises 2,565 tiles in total, including 579 tiles for New York City, 490 for Norfolk, 500 for Washington, D.C., and 996 for Busan across all 12 months, providing a data foundation for subsequent analysis.  To balance the number of samples in different cities, the final dataset contains 23,880 monthly paired samples, including 6,000 samples (500 samples $\times$ 12 months) for New York City, 6,000 for Washington, D.C., 6,000 for Busan, and 5,880 (490 samples $\times$ 12 months) for Norfolk in Appendix Tab.\ref{tab:city_dataset_stats}. 

To support reproducibility and foster further research in urban and energy sustainability, we publicly release the complete MUSE2
dataset on Hugging Face\footnote{\url{https://huggingface.co/datasets/skl24/MUSE2}}. We hope it will serve as a benchmark for the community and advance AI for urban and energy science.






\section{Method}
\label{sec:methodology}

\begin{figure*}
    \centering
    \includegraphics[width=0.9\linewidth]{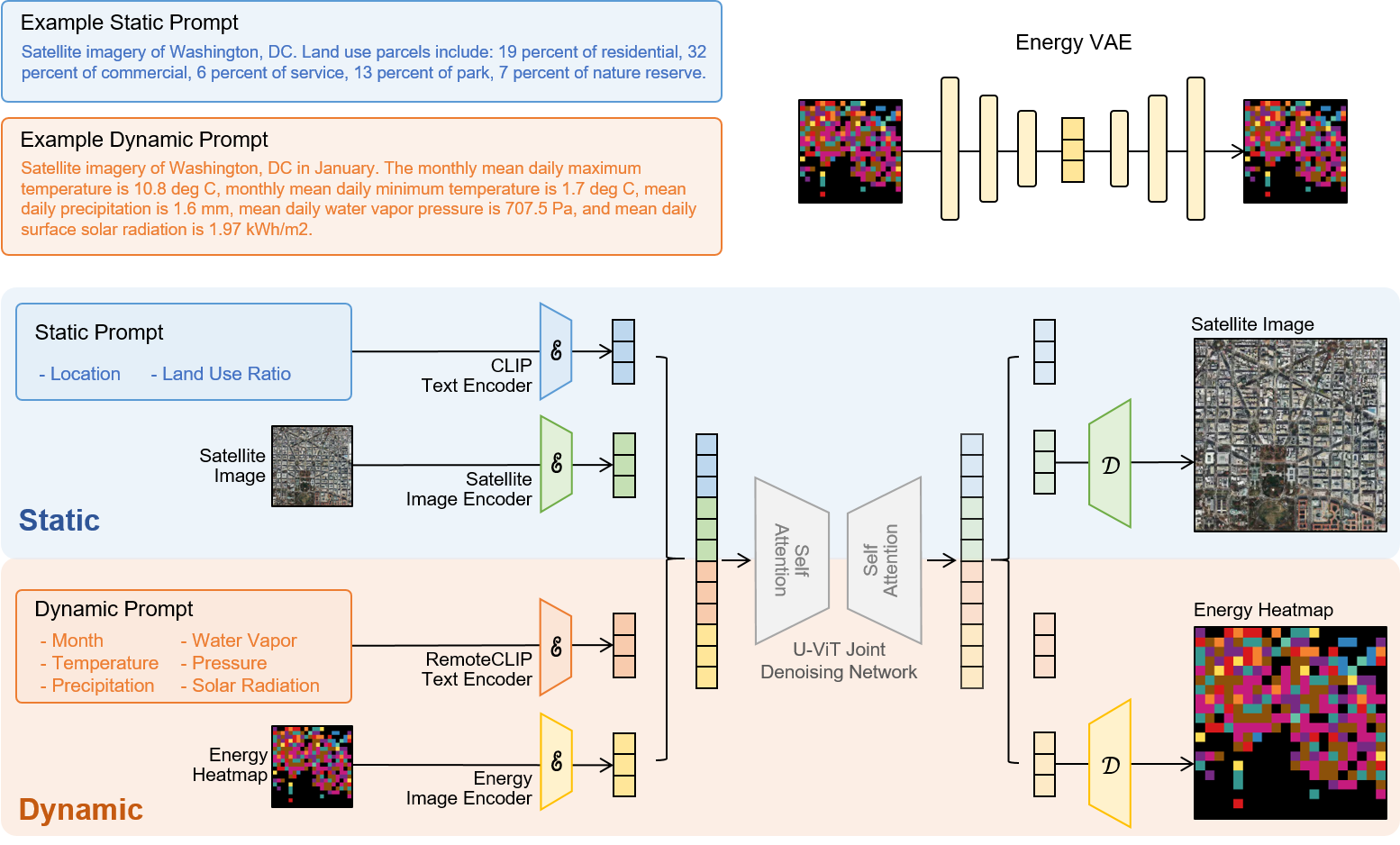}
    \caption{The proposed framework includes four modules: (1) Energy modality-specific variational autoencoders that compress the energy map into compact latent tokens; (2) Two text encoders (CLIP and RemoteCLIP \cite{remoteclip}) for embedding the static and dynamic prompts, respectively; (3) A U-ViT joint denoising network that models the joint distribution of all modalities via the self-attention mechanism; (4) Dynamic monthly energy consumption generation under static urban form.}
    \label{fig:framework}
\end{figure*}

A unified multimodal generative framework is proposed to jointly synthesize realistic static urban satellite imagery $\mathbf{x}$ and the corresponding dynamic monthly building energy consumption map $\mathbf{e}$, conditioned on two urban prompts. In particular, we model the joint distribution:
\begin{equation}
    P(\mathbf{x}, \mathbf{e} \mid \mathbf{c}_{s}, \mathbf{c}_{d}),
\end{equation}
where $\mathbf{c}_{s}$ denotes a static condition that captures the long-term physical urban morphology of a region (location and land-use composition), and $\mathbf{c}_{d}$ denotes a dynamic condition that captures the short-term monthly and climate conditions (the calendar month together with monthly climate variables). We decouple static urban morphology from dynamic monthly energy, enabling the latent diffusion model to envision how the same city consumes energy differently across months and climate conditions in Fig.~\ref{fig:framework}.

\subsection{Multimodal Latent Representations}
Instead of diffusing in high-dimensional pixel space, we use latent representations of image and energy modalities from two individual autoencoders. A pre-trained VAE, with encoder $\mathcal{E}_{x}$ and decoder $\mathcal{D}_{x}$, maps an urban imagery $\mathbf{x}\in\mathbb{R}^{H\times W\times 3}$ to a latent representation $\mathbf{z}_{x}=\mathcal{E}_{x}(\mathbf{x})\in\mathbb{R}^{64\times 64\times 4}$ and reconstructs it via $\hat{\mathbf{x}}=\mathcal{D}_{x}(\mathbf{z}_{x})$.

On the other hand, urban energy consumption maps are spatially sparse, long-tailed, and statistically distinct from RGB imagery. Therefore, we train a domain-specific Energy VAE with encoder $\mathcal{E}_{e}$ and decoder $\mathcal{D}_{e}$ that converts an energy map $\mathbf{e}$ into $\mathbf{z}_{e}=\mathcal{E}_{e}(\mathbf{e})\in\mathbb{R}^{64\times 64\times 4}$ and reconstructs it via $\hat{\mathbf{e}}=\mathcal{D}_{e}(\mathbf{z}_{e})$. In the top-right of Fig.~\ref{fig:framework}, the Energy VAE is trained and optimized with a reconstruction loss on the energy modality, learning a latent spatial representation of the energy consumption.

\subsection{Static and Dynamic Conditioning}
We use two conditioning natural-language prompts and embed them with frozen text encoders. The static prompt encodes the location and the land-use ratios of a region (e.g., ``Satellite imagery of Washington, D.C. Land use parcels include: 19 percent of residential, 32 percent of commercial, 6 percent of service, 13 percent of park, 7 percent of nature reserve.''). It is embedded with a CLIP text encoder to yield the static condition $\mathbf{c}_{s}$ with 77 text-token embeddings.

The dynamic prompt encodes the calendar month and the monthly climate variables, including the mean daily maximum and minimum temperature, precipitation, water vapor pressure, and surface solar radiation (e.g., ``Satellite imagery of Washington, D.C. in January. The monthly mean daily maximum temperature is 10.8 deg C, monthly mean daily minimum temperature is 1.7 deg C, mean daily precipitation is 1.6 mm, mean daily water vapor pressure is 707.5 Pa, and mean daily surface solar radiation is 1.97 kWh/m\textsuperscript{2}.''). Because this monthly and climate information is related to the remote sensing domain \cite{goktepe2025ecomapper}, we embed it with a RemoteCLIP text encoder to obtain the dynamic condition $\mathbf{c}_{d}$ with 77 text-token embeddings. 


\subsection{U-ViT Joint Denoising Network}
Then, we treat these modalities as a joint token sequence and denoise them simultaneously in our framework. To spatially align the generated satellite image with the energy consumption map, we follow U-ViT~\cite{bao2023all} and UniDiffuser~\cite{bao2023one}, jointly diffusing and denoising the two modalities together rather than considering them as two independent processes. Defining $\mathbf{z}_{x}^{0}=\mathbf{z}_{x}$ and $\mathbf{z}_{e}^{0}=\mathbf{z}_{e}$, we achieve this with a single shared timestep $t$ and a single shared Gaussian noise $\boldsymbol{\epsilon}\sim\mathcal{N}(0,\mathbf{I})$:

\begin{equation}
q\!\left(\mathbf{z}_{x}^{t},\mathbf{z}_{e}^{t}\,\middle|\,\mathbf{z}_{x}^{0},\mathbf{z}_{e}^{0}\right)
    = \mathcal{N}\!\left(\sqrt{\alpha_{t}}
    \begin{bmatrix}\mathbf{z}_{x}^{0}\\[2pt] \mathbf{z}_{e}^{0}\end{bmatrix},\,
    (1-\alpha_{t})\mathbf{I}\right),
\end{equation}
where $\alpha_{t}$ is the noise schedule at timestep $t$. Sharing one noise realisation across the stacked latent is useful to learn and model the two modalities' latent relationships in the diffusion process.

The two text prompts serve purely as conditions: we fix their diffusion timestep to $0$ and feed the embeddings $\mathbf{c}_{s}$ and $\mathbf{c}_{d}$ directly into the network. The training loss is:
\begin{equation}
    \mathcal{L} = \mathbb{E}_{t,\,\mathbf{z}_{x}^{0},\,\mathbf{z}_{e}^{0},\,\boldsymbol{\epsilon}}
    \left[\,\big\|\boldsymbol{\epsilon}_{\theta}\!\left(\mathbf{z}_{x}^{t},\mathbf{z}_{e}^{t},\mathbf{c}_{s},\mathbf{c}_{d},t\right)-\boldsymbol{\epsilon}\big\|_{2}^{2}\,\right].
\end{equation}
During training, we perturb the image and energy latents, $\mathbf{z}_{x}^{t}$ and $\mathbf{z}_{e}^{t}$, to learn their joint distribution and cross-modal dependence. At inference, we sample a single Gaussian latent for the joint modality, run the reverse process conditioned on $\mathbf{c}_{s},\mathbf{c}_{d}$, and decode the recovered latents with the urban imagery and energy decoders.


\subsection{Dynamic Monthly Energy Consumption Generation}
In urban science, urban form evolves over long horizons and embodies the cumulative effect of long-term urban development \cite{zhang2026building}, but building energy consumption significantly differs across months and climates \cite{wang2023impacts}. In building science, HVAC systems constitute the single largest end use of building energy \cite{perez2008review,gonzalez2022review}, and space-conditioning loads are the dominant source of seasonal and monthly variation because heating and cooling demand responds directly to outdoor temperature through degree days \cite{american20212021}. 

Thus, holding the urban form fixed, how would monthly dynamic energy consumption respond under different months and climate conditions?

Our framework naturally aligns well with it. As known, a benefit of fitting the joint distribution $q(\mathbf{z}_{x}^{0}, \mathbf{z}_{e}^{0})$~\cite{bao2023one} is that the conditional distribution $q(\mathbf{z}_{e}^{0}\mid\mathbf{z}_{x}^{0})$ can be inferred without model retraining. Thus, at inference, we then fix the urban imagery latent token $\mathbf{z}_{x}^{0}$ and progressively denoise the energy latent token at every step:


\begin{equation}
    p_{\theta}\!\left(\mathbf{z}_{e}^{t-1}\mid\mathbf{z}_{e}^{t},\mathbf{z}_{x}^{0},\mathbf{c}_{s},\mathbf{c}_{d}\right)
    =\mathcal{N}\!\left(\boldsymbol{\mu}_{\theta}\!\left(\mathbf{z}_{e}^{t},\mathbf{z}_{x}^{0},\mathbf{c}_{s},\mathbf{c}_{d},t\right),\,\sigma_{t}^{2}\mathbf{I}\right).
\end{equation}
Consequently, fixing the static prompt $\mathbf{c}_{s}$ and urban imagery latent token $\mathbf{z}_{x}^{0}$ in the latent space, while changing the dynamic prompt $\mathbf{c}_{d}$ (e.g., switching the calendar month or the temperature and solar-radiation profile) generates the monthly and seasonal energy consumption under the same urban imagery. It aligns monthly changes in dynamic energy consumption under the static urban form. This enables what-if analysis for planners to envision how a city's energy consumption dynamically changes across months and climate conditions, further supporting peak-load analysis and demand response.

\section{Experiments}
In this section, we conduct extensive experiments to answer the following research questions:
\begin{itemize}[leftmargin=*]
    \item \textbf{RQ1:} Can the proposed framework jointly synthesize urban imagery together with spatially aligned energy consumption maps?
    \item \textbf{RQ2:} How effectively does the proposed framework generate monthly dynamic building energy consumption maps with static urban morphology? 
    \item \textbf{RQ3:} To what extent does DySENSE improve state-of-the-art urban building energy prediction models with limited data?

\end{itemize}

\subsection{Experimental Setups}
All experiments were run on a 64-bit Ubuntu 24.04 platform with an AMD EPYC 9754 128-Core Processor, 1.5TB of RAM, and 10 NVIDIA L40S GPUs (48\,GB each) under CUDA 13.1. We implemented our framework in Python 3.11 with PyTorch 2.11. The total training time is more than 800 GPU hours. The training time of the main generative AI model is about 220 GPU hours while the training time of the downstream prediction models is more than 600 GPU hours. Also, the sampling/inference time of the main generative AI model is about 60 GPU hours. Full implementation details and hyperparameters are reported in Appendix~\ref{sec:Implementation}.

\begin{figure*}
    \centering
    \includegraphics[width=1\linewidth]{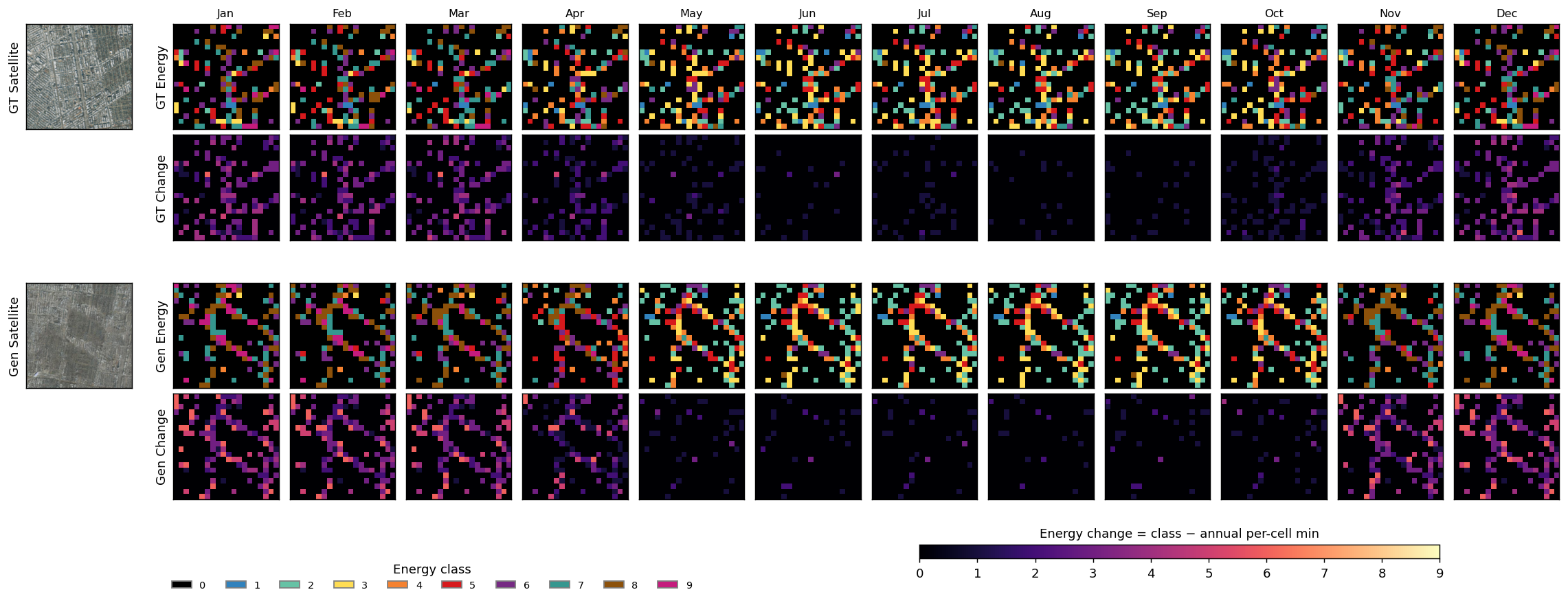}
    \caption{Qualitative visualization of ground-truth and generated urban imagery and monthly energy consumption. The Energy changes are computed by the difference between monthly energy consumption values and the annual minimum values. Our generated monthly energy consumption and its trend are very similar to the ground truth.}
    \label{fig:res-monthly}
\end{figure*}

\subsection{Energy Metric}
Semantic segmentation metrics (e.g., mIoU) assess pixel-level accuracy, yet urban energy planners are primarily concerned with the reliability of aggregate energy demand. We therefore adopt the industry-standard calibration metrics specified in ASHRAE Guideline 14 \cite{american2014ashrae,ROYAPOOR2015109}: the Normalized Mean Bias Error (NMBE) and the Coefficient of Variation of the Root Mean Square Error (CVRMSE). The details can be found in Appendix~\ref{energy-me}.



\subsection{Static Generative Performance Evaluation}\label{gpe}

After training DySENSE, we evaluate the quality of the generated satellite imagery and building energy consumption maps.

\subsubsection{Quantitative Evaluation}

We quantitatively assess the generated satellite imagery using Fréchet Inception Distance (FID), Peak Signal-to-Noise Ratio (PSNR), and Learned Perceptual Image Patch Similarity (LPIPS). The proposed framework achieves an FID of 36.7362, a reasonable distribution similarity between generated and real satellite images. The LPIPS score of 0.5604 suggests that the generated images preserve perceptual characteristics of real urban images. DySENSE achieves a PSNR of 11.5177, indicating that the generated images preserve a certain degree of pixel-level fidelity. Overall, results show that DySENSE successfully encodes complex urban morphologies, and can generate compliant, realistic, and visually plausible urban imagery. 

Beyond generated urban imagery, it is important to evaluate whether the generated energy consumption map aligns with the generated urban imagery. A good generative framework should yield urban imagery-energy pairs that are not only visually realistic \cite{sun2026envisioning} but also spatially aligned between urban form and energy consumption. It can be conducted by training the downstream prediction task \cite{wu2023datasetdm,wang2025jodiffusion}. Thus, we quantitatively assess the alignment of the generated satellite imagery and energy consumption map by comparing the downstream energy prediction performance. 
Specifically, we evaluate different training strategies for the downstream energy prediction task: \textbf{(A)  Real Only (Baseline):} The prediction model is trained on the limited available real data with augmentation.  \textbf{ (B1) Static Mix Training (Ours):} The prediction model is trained on the same limited real data augmented with a large corpus of synthetic static imagery-energy samples. 
We benchmark several SOTA prediction models on MUSE2. To prevent data leakage, we train both our generative framework and downstream prediction models on the same real training data for each ratio. For the testing set, we fix its size at 20\%, which is split by tile to prevent data leakage. And all prediction models were evaluated on this fixed real-world test set to ensure a fair comparison.

We have some findings. In Tab.~\ref{tab:energy-overall-energy}, compared with SOTA urban building energy prediction methods~\citep{yap2025revealing, MAYER2023120542, sun2026deciphering}, our generative data augmentation with the Static Mix strategy improves energy prediction performance across all methods on the
limited data. These gains are consistent across predictors, improving every baseline metric. In particular, our Static Mix strategy improves energy prediction performance by about 6.8\% in mIoU and substantially reduces energy prediction error, lowering $|$NMBE$|$ by about 3.7\% and CVRMSE by 16.7\%.

Fig.~\ref{fig-combined-metrics} further reveals how these gains scale with the
increase in real data. (1) Across all strategies, energy prediction performance
improves monotonically as the training-set ratio increases.
(2) On the limited real data, the Static Mix strategy yields a substantial gain over
the real-only baseline (e.g., $+7.0$\,mIoU at the $6.25\%$ ratio and $+6.4$\,mIoU
at the $12.5\%$ ratio, with CVRMSE reduced from $82.3\%$ to $58.4\%$ at
$6.25\%$; Fig.~\ref{fig-combined-metrics}~c,d). (3) This advantage narrows as more real data becomes available in the full-data setting, demonstrating that the synthetic imagery-energy pair is significantly effective when real data is limited. Together, these results provide evidence that the generated imagery and building energy maps are spatially aligned.

\subsubsection{Qualitative Evaluation}

\begin{figure*}[ht]
    \centering
    \includegraphics[width=0.85\linewidth]{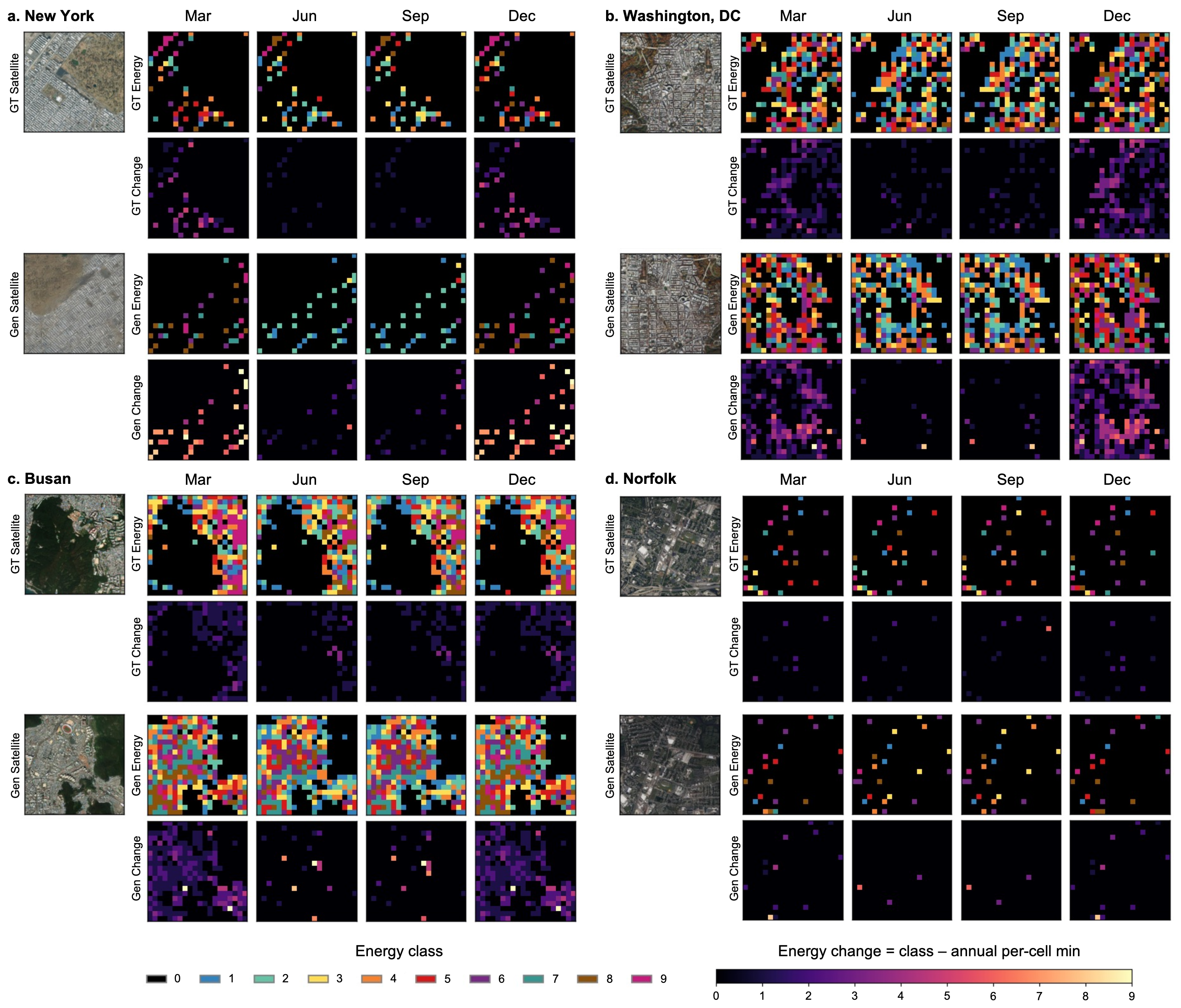}
    \caption{Qualitative visualization of generated results across 4 cities.  We sample some monthly generated energy results across seasons. (a) The comparison between GT and generated urban imagery and monthly building energy consumption maps in NYC. (b,c,d) Other cases in Washington, D.C., Busan, and Norfolk. The dynamic energy consumption trends are similar. }
    \label{fig:res-cases}
\end{figure*}

Fig. \ref{fig:res-monthly} and Fig. \ref{fig:res-cases} provide the visualisation of the generated samples across four distinct metropolitan areas: NYC, Washington, D.C., Norfolk, and Busan. They confirm that the generated imagery and energy maps are visually and spatially well aligned.

\subsection{Dynamic Generative Performance Evaluation}
\label{dpe}

\begin{table*}
    \centering
    \caption{12-month energy prediction performance comparison on the fixed real test set (20\%) using ASHRAE metrics \cite{american2014ashrae}. Strategy B uses real + synthetic training data; Strategy A1 uses real data only; Strategy A2 uses real data with equal-size simple duplication; Strategy A3 uses real data with equal-size simple augmentation by Gaussian noise. NMBE is reported with their original signs; values closer to 0 indicate better performance. More results can be found in Appendix Tab. \ref{tab:per_class_iou_acc}.}
    \label{tab:energy-overall-energy}
    \setlength{\tabcolsep}{15pt}
    \begin{tabular}{ll cc cc}
        \toprule
        Method & Training Data & Acc$\uparrow$ & mIoU$\uparrow$ & NMBE & CVRMSE $\downarrow$ \\
        \midrule
        \citet{yap2025revealing} & Strategy A1: Real & 0.5343 & 0.3687 & 6.02\% & 33.58\% \\
        \citet{yap2025revealing} & Strategy A2: Real duplication  &	0.5340	&0.3686	&7.73\%	&32.56\%\\
        \citet{yap2025revealing} & Strategy A3: Real augmentation  &	0.5462	&0.3712	&5.57\%	&30.70\%\\
        \citet{yap2025revealing} + \textbf{Ours} &\textbf{Strategy B1: Static Mix} & \textbf{0.6034} & \textbf{0.4364} & \textbf{3.37\%} & \textbf{16.91\%} \\
        \citet{yap2025revealing} + \textbf{Ours} & \textbf{Strategy B2: Dynamic Mix} & \textbf{0.6217} & \textbf{0.4555} & \textbf{2.82\%} & \textbf{15.31\%} \\
        \text{Diff } ( |\text{Ours}| - |\text{Baseline}| ) & & +0.0874 & +0.0868 & -3.20\% & -18.27\% \\
        \midrule
        \citet{MAYER2023120542} & Strategy A1: Real & 0.5887 & 0.4212 & 3.83\% & 24.27\% \\
        \citet{MAYER2023120542} & Strategy A2: Real duplication & 0.5887	&0.4212	&3.96\%	&25.78\%\\        
        \citet{MAYER2023120542} & Strategy A3: Real augmentation & 0.5932	&0.4263	&4.01\%	&20.65\%\\
        \citet{MAYER2023120542} + \textbf{Ours} & \textbf{Strategy B1: Static Mix} & \textbf{0.6163} & \textbf{0.4497} & \textbf{3.44\%} & \textbf{16.86\%} \\
        \citet{MAYER2023120542} + \textbf{Ours} & \textbf{Strategy B2: Dynamic Mix} & \textbf{0.6274} & \textbf{0.461} & \textbf{2.87\%} & \textbf{15.26\%} \\
        \text{Diff } ( |\text{Ours}| - |\text{Baseline}| ) & & +0.0387 & +0.0398 & -0.96\% & -9.01\% \\
        \midrule
        \citet{sun2026deciphering} & Strategy A1: Real & 0.5372 & 0.3708 & 7.53\% & 30.57\% \\
        \citet{sun2026deciphering} & Strategy A2: Real duplication &  0.5339& 	0.3694& 	6.95\%	& 28.49\%\\
        \citet{sun2026deciphering} & Strategy A3: Real augmentation &  0.5540& 	0.3894& 	6.60\%	& 24.90\%\\
        \citet{sun2026deciphering} + \textbf{Ours} & \textbf{Strategy B1: Static Mix} & \textbf{0.5969} & \textbf{0.4293} & \textbf{3.86\%} & \textbf{19.40\%} \\
        \citet{sun2026deciphering} + \textbf{Ours} & \textbf{Strategy B2: Dynamic Mix} & \textbf{0.5999} & \textbf{0.4327} & \textbf{3.42\%} & \textbf{16.62\%} \\
        \text{Diff } ( |\text{Ours}| - |\text{Baseline}| ) & & +0.0627 & +0.0619 & -4.11\% & -13.95\% \\
        \bottomrule
    \end{tabular}
    \vspace{-1em}
\end{table*}

In practice, monthly real-world energy data are more scarce and limited. Thus, we generated monthly aligned urban energy data using our framework with the proposed monthly energy consumption generation method.






\subsubsection{Quantitative Evaluation}
We quantitatively assess the distributional fidelity between the generated monthly energy consumption and the ground-truth monthly energy consumption. The generated monthly energy consumption maps closely match the real energy distribution, with low KL divergence (0.0024), JS divergence (0.0006), and TV distance (0.0181). Even after removing the dominant background class 0, the model maintains good foreground energy consistency (KL = 0.0085, JS = 0.0021, TV = 0.0496), indicating that DySENSE effectively captures the monthly distribution pattern of the ground-truth dynamic energy consumption.


Then, to address RQ2 and RQ3, we conducted data augmentation experiments on downstream tasks for urban building energy consumption prediction.  To achieve this, we utilise a multimodal prediction model in which the inputs are static satellite imagery and dynamic information (including month, weather, etc.), and the output is an energy consumption map.  We add another training strategy for the downstream energy prediction task: \textbf{(B2) Dynamic Mix Training (Ours):} The prediction model is trained on the same limited real data augmented with a large corpus of synthetic samples (imagery and dynamic monthly energy maps).

\begin{figure*}[t]
    \centering
    \includegraphics[width=0.86\linewidth]{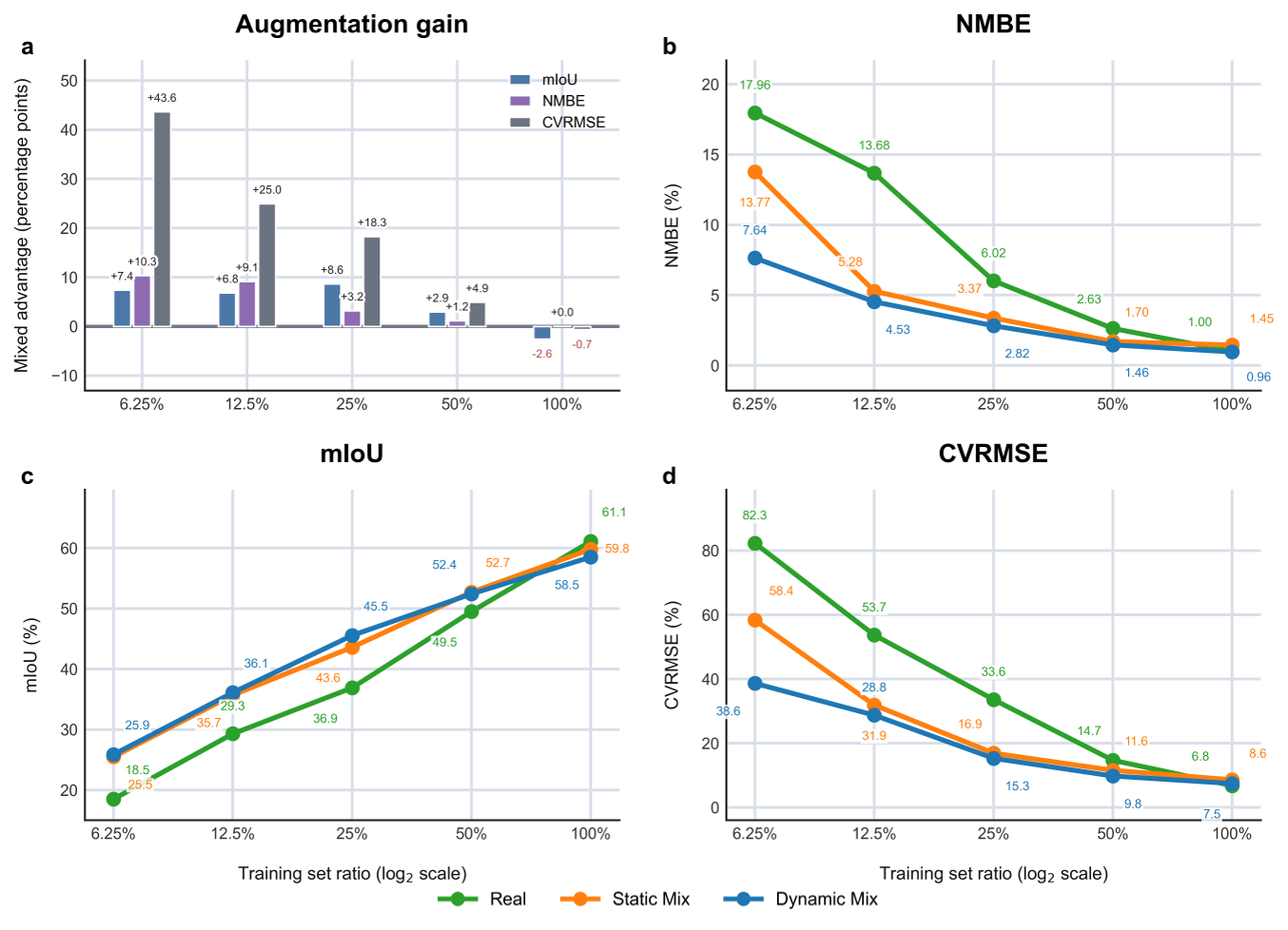}
    \caption{Generative augmentation effect on downstream building energy prediction metrics across different training set ratios.}
    \label{fig-combined-metrics}
\end{figure*}

Tab~\ref{tab:energy-overall-energy} reports the quantitative comparison against
SOTA methods. We have some findings. (1) Adding our generated paired urban imagery and monthly energy data consistently improves the downstream building energy prediction of all baselines. (2) Under the Dynamic Mix strategy, mIoU improves by about 8.68\% and CVRMSE reduces by 9\%-18\%. For \citet{yap2025revealing}, which corresponds to the $25\%$ training data in Fig.~\ref{fig-combined-metrics}, our Dynamic Mix raises accuracy from $0.5343$ to $0.6217$ (8.7\%) and mIoU from $0.3687$ to $0.4555$ (8.7\%), while reducing NMBE from $6.02\%$ to $2.82\%$ and CVRMSE from $33.58\%$ to $15.31\%$ (18.3\%). (3) These improvements align with the trends in Fig.~\ref{fig-combined-metrics}, i.e., our framework significantly enhances urban monthly energy prediction under limited data.

Fig.~\ref{fig:res-monthly} shows the full 12-month sequence for a tile, where the generated energy maps and their month-to-month changes closely track the ground truth, capturing monthly peaks and trends. Fig.~\ref{fig:res-cases} extends this across cities, demonstrating that the generated dynamic monthly energy trends are consistent with ground truth in each city. In particular, in most cities, winter energy consumption is higher than in other seasons because heating systems often consume more energy to keep indoor spaces warm in winter.

Besides, in Appendix Fig.~\ref{fig:3city}, DySENSE successfully captures the monthly energy consumption patterns across different cities. The generated curves follow the same monthly trends as the ground truth, with a high average Pearson correlation of 0.96 (e.g., 0.93 for Busan and 0.98 for New York City), indicating that DySENSE can preserve seasonal and monthly energy dynamics across all cities.

\subsection{Ablation Experiment}

We conducted experiments to evaluate the selection of RemoteCLIP and CLIP in Appendix Section \ref{sec:clip} and Tabs. \ref{tab:ablation_gen_text_encoder} and \ref{tab:ablation_text_encoder}. We find that RemoteCLIP consistently yields better performance on both generative and predictive tasks.  We think this is because the pretrained RemoteCLIP has learned image-text information in remote sensing domain \cite{remoteclip} and thus is better suited to this study.


\section{Discussion}
\subsection{Domain Contributions and Impacts} 

As rapid urbanisation reshapes global cities, effective modelling and planning of urban energy dynamics are important for policy-making and peak-load assessment, advancing the UN's SDGs 7 and 11. Our key contribution lies in transforming the existing UBEM from a post-design assessment into an active, generative, energy-efficient urban planning process. Instead of predicting energy consumption for an existing city or static annual energy plan \cite{sun2026sense}, DySENSE allows planners to envision diverse future cities that do not yet exist across design variables (e.g., land use and climate) and immediately capture dynamic temporal urban energy consumption patterns, supporting peak-load analysis and management, building performance simulation, and high-resolution temporal dynamic demand response applications for urban and energy science.


One key data challenge in data-driven UBEM is the scarcity and confidentiality of fine-grained temporal monthly energy data because of privacy, cost, sensitivity, etc \cite{zhou2025state}. We mitigate it by synthesizing high-resolution urban imagery and dynamic monthly energy data aligned with real urban morphology for open benchmarking (Tab~\ref{tab:energy-overall-energy}). Trained on only 25\% of the energy data, DySENSE generates virtually unlimited annotated urban imagery and monthly energy data, significantly improving the performance of SOTA urban building energy prediction methods. One challenge in AI methods is the inability of existing diffusion models to generate urban functional dynamics, which we address by training Energy VAE, spatially denoising the joint latent representation from urban imagery and fine-grained urban energy, temporally decoupling static urban morphology from dynamic energy, and generating monthly energy conditioned on the imagery latent.

In summary, DySENSE can offer government agencies, urban and energy scientists, and developers a powerful generative tool for envisioning future urban scenarios and temporal dynamics of urban energy consumption for peak-load analysis, energy-efficient planning and retrofits, and energy demand response optimisation. By releasing the MUSE2 dataset, DySENSE codes and trained models as open access, we advocate for a decarbonised, energy-efficient planning process aligned with SDGs 7 and 11, advancing the development of AI for urban science, energy systems and building science.

\subsection{Limitations and Ethical Considerations}


While DySENSE generalises well across four metropolitan areas in America and Asia, its geographic coverage remains limited. Future work can extend the insight to more cities, climate zones, and urban morphologies, especially in data-scarce developing regions. Future work could integrate external validation (e.g., utility audit or smart meter data) to improve broader real-world generalizability. Future studies can explore the proposed framework for building operational carbon emissions by jointly considering energy consumption sources, spatial and temporal grid carbon intensity, and emission factors \cite{yap2025revealing,qiubetter2018}.

All data in MUSE2 come from publicly available, openly licensed sources, e.g., OpenStreetMap, Mapbox, municipal energy disclosure records (e.g., NYC Local Law 84), and public administrative data from the Korean government—with no private or household-level data, with no use of private or non-consensual data.



\section{Conclusion}
This study presents DySENSE, a joint generative framework that synthesizes static urban satellite imagery together with spatially aligned, dynamic monthly urban energy maps. On the four-cities dataset, the generated monthly energy closely matches real distributions, and augmenting scarce real-world data consistently improves three SOTA predictors. We hope our dataset and framework can facilitate scientific development in AI for urban science, energy systems, and building science toward SDGs 7 and 11.




\section{Generative AI Usage}

The authors declare that AI tools (e.g., Codex) were used solely for language polishing and code assistance. All scientific contributions, including ideas, experiments, and analyses, are the authors' own. All intellectual content, data analysis, interpretations, and conclusions were conceived, written, and verified by the authors. 

\bibliographystyle{ACM-Reference-Format}
\bibliography{sample-base}

\appendix

\section{Appendix}
\subsection{Implementation Details}
\label{sec:Implementation}

The hyperparameters are reported in Tabs. \ref{tab:hyper_vae} and \ref{tab:hyper_ldm}.
\begin{table}[!h]
\centering
\caption{Training hyperparameters of the Energy VAE.}
\label{tab:hyper_vae}
\begin{tabular}{ll}
\toprule
\textbf{Hyperparameter} & \textbf{4 cities, 10-class} \\
\midrule
in channels & 4 \\
intermediate channels & 512 \\
out channels & 10 \\
block out channels & (128, 256, 512, 512) \\
resolution & 512 \\
training augmentation & \makecell{Resize, RandomCrop, \\ RandomHorizontalFlip} \\
batch size & 96 \\
epoch & 30 \\
optimizer & AdamW \\
learning rate & 1e-5 \\
AdamW - $\epsilon$ & 1e-8 \\
AdamW - weight decay & 5e-2 \\
\bottomrule
\end{tabular}
\end{table}

\begin{table}[!h]
\centering
\caption{Training hyperparameters of the joint diffusion.}
\label{tab:hyper_ldm}
\begin{tabular}{ll}
\toprule
\textbf{Hyperparameter} & \textbf{4 cities, 10-class} \\
\midrule
resolution & 512 \\
sequence length & 2206 \\
batch size & 12  \\
epoch & 80 \\
learning rate & 5e-5 \\
lr scheduler & cosine (500 warmup steps) \\
AdamW - $\epsilon$ & 1e-8 \\
AdamW - weight decay & 1e-2 \\
\bottomrule
\end{tabular}
\end{table}

\subsection{Energy Metric}\label{energy-me}

NMBE captures the systematic bias of the estimates, whereas CVRMSE characterizes the dispersion of the prediction errors. We recover the physical energy consumption values from the discrete class predictions via the expm1 transform, and compute both metrics as

\begin{equation}
    \text{NMBE} = \frac{\sum_{i=1}^{N} (\hat{y}_i - y_i)}{\sum_{i=1}^{N} y_i} \times 100\%,
\end{equation}

\begin{equation}
    \text{CVRMSE} = \frac{\sqrt{\frac{1}{N} \sum_{i=1}^{N} (\hat{y}_i - y_i)^2}}{\bar{y}} \times 100\%,
\end{equation}
where $y_i$ and $\hat{y}_i$ denote the ground-truth and predicted energy of the $i$-th tile, respectively, $\bar{y}$ is the mean of the ground-truth values, and $N$ is the number of test samples.


\subsection{Generation energy performance analysis with ground truth}
\label{sec:3city}

\begin{figure*}[!h]
    \centering
    \includegraphics[width=0.9\linewidth]{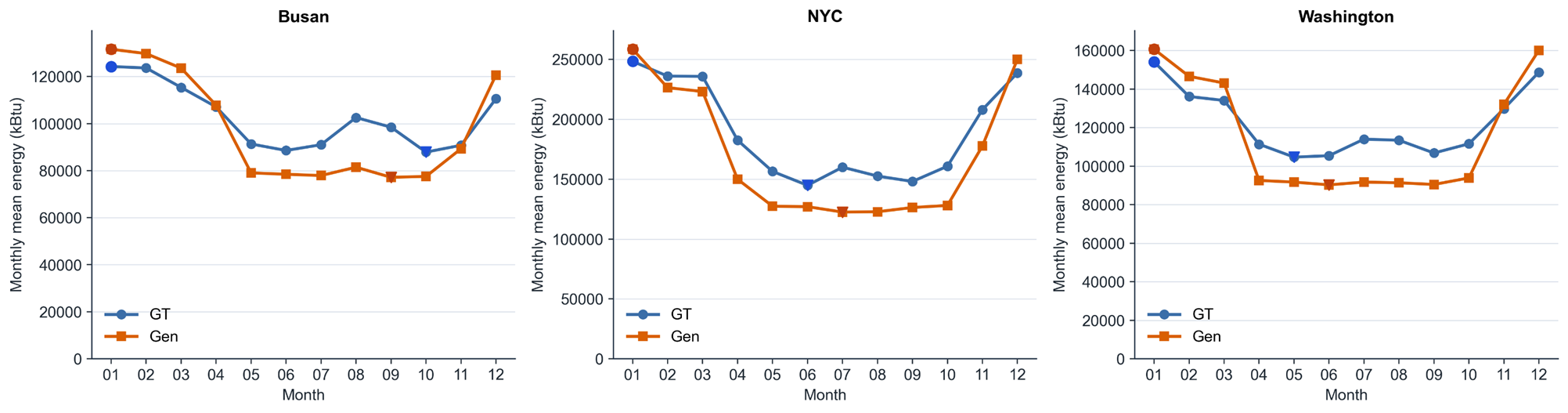}
    \caption{Monthly energy consumption patterns generated by DySENSE across different cities. The curves compare the monthly mean energy consumption of generated maps and ground-truth maps. DySENSE captures seasonal and monthly dynamics well, with peak months exactly aligned with the ground truth in different cities.}
    \label{fig:3city}
\end{figure*}

In Fig.~\ref{fig:3city}, DySENSE successfully captures the monthly energy consumption patterns across different cities. The peak month aligns exactly with the ground truth across cities, and the U-shaped seasonal cycle is well reproduced. In Busan, the framework tends to overestimate the winter peak, whereas in NYC and Washington, D.C., it slightly underestimates the summer peak. These results indicate that DySENSE preserves the dominant seasonal and monthly energy dynamics across cities.

\subsection{Ablation experiment: RemoteCLIP v.s. CLIP}
\label{sec:clip}
We conduct an ablation experiment in which CLIP replaces the RemoteCLIP text encoder in the proposed generative framework. In Tab. \ref{tab:ablation_gen_text_encoder}, we replace the
RemoteCLIP text encoder with the vanilla CLIP ViT-L-14 text encoder, with the land-use CLIP encoder and all hyperparameters unchanged. We find that the RemoteCLIP consistently yields better image quality across all metrics.  We think this is because the pretrained RemoteCLIP has learned remote sensing-related text information \cite{remoteclip} and is better suited to this satellite imagery generative task than CLIP.

We also conduct another ablation experiment for the energy consumption prediction tasks. In Tab. \ref{tab:ablation_text_encoder}, we find that the overall performance of RemoteCLIP is higher than that of CLIP. We think this is because RemoteCLIP has learned remote sensing-related text well and is better suited than CLIP for this remote sensing task.

\begin{table}[!h]
\centering
\caption{Ablation study of the climate text encoder in the proposed generative framework.}
\label{tab:ablation_gen_text_encoder}
\begin{tabular}{lcc}
\toprule
Metric & RemoteCLIP & CLIP \\
\midrule
FID\,$\downarrow$   & \textbf{36.7362} & 37.4800 \\
PSNR\,$\uparrow$    & \textbf{11.5177} & 10.9076 \\
LPIPS\,$\downarrow$ & \textbf{0.5604}  & 0.5695  \\
\bottomrule
\vspace{-1em}
\end{tabular}
\end{table}

\begin{table}[ht]
\centering
\caption{Ablation on the dynamic climate text encoder of the downstream prediction model:
RemoteCLIP vs.\ CLIP.}
\label{tab:ablation_text_encoder}
\footnotesize
\setlength{\tabcolsep}{3.5pt}
\resizebox{\columnwidth}{!}{%
\begin{tabular}{lcccccc}
\toprule
\multirow{2}{*}{Ratio}
  & \multicolumn{2}{c}{mIoU\,$\uparrow$}
  & \multicolumn{2}{c}{$|$NMBE$|$ (\%)\,$\downarrow$}
  & \multicolumn{2}{c}{CVRMSE (\%)\,$\downarrow$} \\
\cmidrule(lr){2-3} \cmidrule(lr){4-5} \cmidrule(lr){6-7}
 & RemoteCLIP & CLIP & RemoteCLIP & CLIP & RemoteCLIP & CLIP \\
\midrule
\multicolumn{7}{l}{\textit{Strategy A: Real Only}} \\
1/16 & 18.50 & \textbf{18.92} & \textbf{17.96} & 28.93 & \textbf{82.30} & 103.12 \\
1/8  & 29.30 & \textbf{29.36} & \textbf{13.68} & 16.99 & \textbf{53.70} &  61.96 \\
1/4  & \textbf{36.90} & 35.68 &  \textbf{6.02} &  7.32 & 33.60 &  \textbf{33.43} \\
1/2  & \textbf{49.50} & 48.59 &  \textbf{2.63} &  3.18 & 14.70 &  \textbf{13.13} \\
full & \textbf{61.10} & 60.42 &  \textbf{1.00} &  1.37 &  \textbf{6.80} &   8.06 \\
\midrule
\multicolumn{7}{l}{\textit{Strategy B2: Dynamic Mix (Ours)}} \\
1/16 & 25.87 & \textbf{26.76} &  \textbf{7.64} &  9.80 & \textbf{38.65} &  57.08 \\
1/8  & 36.11 & \textbf{36.16} &  4.53 &  \textbf{4.18} & 28.75 &  \textbf{24.37} \\
1/4  & \textbf{45.55} & 44.82 &  \textbf{2.82} &  3.56 & \textbf{15.31} &  16.88 \\
1/2  & \textbf{52.42} & 51.50 &  \textbf{1.46} &  2.03 &  \textbf{9.80} &  11.47 \\
full & \textbf{58.51} & 58.30 &  \textbf{0.96} &  1.44 &  \textbf{7.46} &   8.60 \\
\bottomrule
\end{tabular}%
}
\end{table}





\begin{table*}
\centering
\caption{City-level statistics data in MUSE2. Each tile has 12-month energy records.}
\label{tab:city_dataset_stats}
\footnotesize
\setlength{\tabcolsep}{5pt}
\begin{tabular}{lccccc}
\toprule
City & Satellite imagery & Urban density &Original energy map  & Filtered energy map & Used energy map \\
\midrule
New York City & 1589& 1589 & 1589& 579 & 500
 \\
 Busan & 1438& 1438 & 1438& 996 & 500 \\
Washington, D.C.& 1914& 1914 & 1914& 500 & 500
 \\
Norfolk & 1328& 1328 & 1328& 490  & 490
 \\
\bottomrule
\end{tabular}
\end{table*}


\begin{table*}
\centering
\caption{Per-class IoU and accuracy results under different training strategies.}
\label{tab:per_class_iou_acc}
\scriptsize
\setlength{\tabcolsep}{2.2pt}
\resizebox{\textwidth}{!}{%
\begin{tabular}{ll*{18}{c}}
\toprule
\multirow{2}{*}{Method} & \multirow{2}{*}{Training Data}
& \multicolumn{9}{c}{IoU (per class)}
& \multicolumn{9}{c}{Acc (per class)} \\
\cmidrule(lr){3-11} \cmidrule(lr){12-20}
& & C1 & C2 & C3 & C4 & C5 & C6 & C7 & C8 & C9
  & C1 & C2 & C3 & C4 & C5 & C6 & C7 & C8 & C9 \\
\midrule
\citet{yap2025revealing} & Strategy A: real
& 0.5106 & 0.3666 & 0.3188 & 0.2930 & 0.3059 & 0.2938 & 0.3121 & 0.3684 & 0.5488
& 0.6731 & 0.5189 & 0.4780 & 0.4395 & 0.5033 & 0.4353 & 0.4495 & 0.5643 & 0.7470 \\
\citet{yap2025revealing} + Ours & Strategy B1: Static Mix
& 0.5790 & 0.4304 & 0.3857 & 0.3706 & 0.3609 & 0.3637 & 0.3873 & 0.4319 & 0.6179
& 0.7350 & 0.5899 & 0.5375 & 0.5676 & 0.5165 & 0.5195 & 0.5596 & 0.6141 & 0.7903 \\
\citet{yap2025revealing} + Ours & Strategy B2: Dynamic Mix
& 0.5911 & 0.4544 & 0.3997 & 0.3821 & 0.3857 & 0.3927 & 0.4104 & 0.4545 & 0.6285
& 0.7278 & 0.6303 & 0.5589 & 0.5485 & 0.5481 & 0.5759 & 0.5844 & 0.6229 & 0.7983 \\
\midrule
\citet{MAYER2023120542}& Strategy A: real
& 0.5592 & 0.4261 & 0.3735 & 0.3463 & 0.3541 & 0.3576 & 0.3710 & 0.4164 & 0.5866
& 0.6988 & 0.6127 & 0.5509 & 0.5066 & 0.5192 & 0.5169 & 0.5236 & 0.6041 & 0.7649 \\
\citet{MAYER2023120542} + Ours & Strategy B1: Static Mix
& 0.5925 & 0.4496 & 0.4016 & 0.3773 & 0.3791 & 0.3824 & 0.3976 & 0.4447 & 0.6225
& 0.7323 & 0.6179 & 0.5737 & 0.5403 & 0.5456 & 0.5608 & 0.5543 & 0.6184 & 0.8036 \\
\citet{MAYER2023120542} + Ours & Strategy B2: Dynamic Mix
& 0.5985 & 0.4562 & 0.4098 & 0.3905 & 0.3926 & 0.3996 & 0.4115 & 0.4602 & 0.6304
& 0.7435 & 0.6188 & 0.5785 & 0.5498 & 0.5594 & 0.5735 & 0.5775 & 0.6442 & 0.8013 \\
\midrule
\citet{sun2026deciphering} & Strategy A: real
& 0.5108 & 0.3674 & 0.3241 & 0.2870 & 0.3024 & 0.3058 & 0.3177 & 0.3630 & 0.5587
& 0.6767 & 0.5201 & 0.4874 & 0.4094 & 0.4647 & 0.4690 & 0.4747 & 0.5612 & 0.7716 \\
\citet{sun2026deciphering}+ Ours & Strategy B1: Static Mix
& 0.5729 & 0.4323 & 0.3797 & 0.3595 & 0.3583 & 0.3600 & 0.3782 & 0.4215 & 0.6010
& 0.6990 & 0.5924 & 0.5290 & 0.5262 & 0.5153 & 0.5298 & 0.5452 & 0.6209 & 0.8143 \\
\citet{sun2026deciphering}+ Ours & Strategy B2: Dynamic Mix
& 0.5666 & 0.4264 & 0.3820 & 0.3644 & 0.3618 & 0.3680 & 0.3845 & 0.4316 & 0.6089
& 0.7096 & 0.5860 & 0.5428 & 0.5325 & 0.5158 & 0.5453 & 0.5570 & 0.6204 & 0.7898 \\
\bottomrule
\end{tabular}%
}
\end{table*}





\begin{figure}
    \centering
    \includegraphics[width=0.9\linewidth]{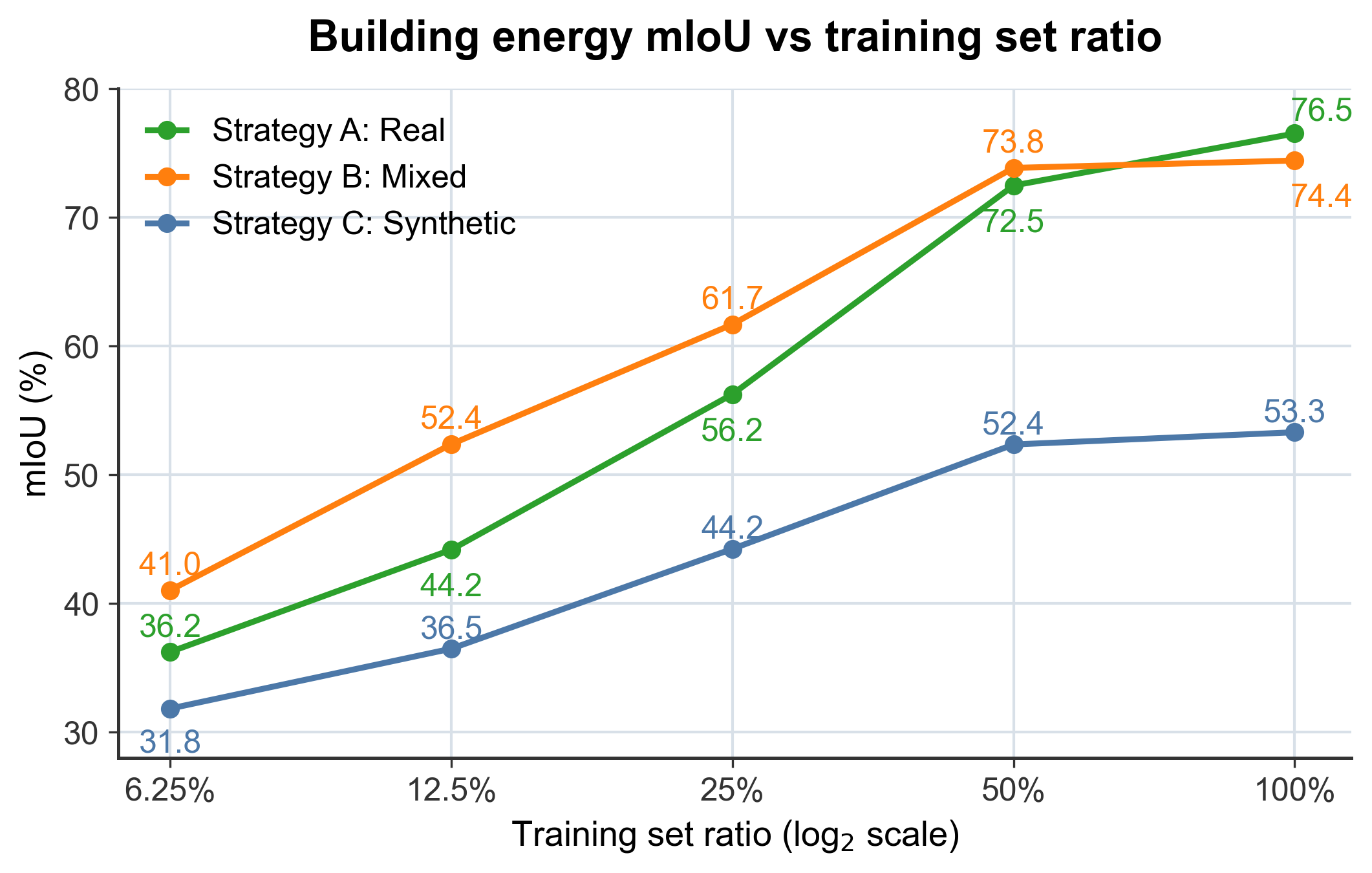}
    \caption{Performance comparison on the downstream building energy consumption prediction task under different training-data strategies.}
    \label{fig:downstream_charts}
\end{figure}



\subsection{Physical energy consumption Recovery}


Each class is mapped to a documented representative value, and then the continuous physical energy consumption values (unit:kBtu) are recovered via expm1. The resulting representative values on the original kBtu scale are reported in Tab~\ref{tab:representative-energy}. The continuous energy-error metrics (NMBE/CVRMSE) were already evaluated on the original kBtu scale. 

\begin{table}[!b]
\centering
\caption{Representative continuous energy-consumption values for each class.}
\setlength{\tabcolsep}{2pt}
\label{tab:representative-energy}
\footnotesize
\begin{tabular}{c r r r r r r r r}
\hline
Class
& \multicolumn{2}{c}{Busan}
& \multicolumn{2}{c}{New York City}
& \multicolumn{2}{c}{Norfolk}
& \multicolumn{2}{c}{Washington}
\\

\cline{2-3}
\cline{4-5}
\cline{6-7}
\cline{8-9}

& kBtu & $\mathrm{log1p}$
& kBtu & $\mathrm{log1p}$
& kBtu & $\mathrm{log1p}$
& kBtu & $\mathrm{log1p}$
\\

\hline
0 & 0.00       & 0.000
  & 0.00       & 0.000
  & 0.00       & 0.000
  & 0.00       & 0.000 \\

1 & 1693.45    & 7.435
  & 42538.34   & 10.658
  & 229.11     & 5.439
  & 63285.79   & 11.055 \\

2 & 7777.19    & 8.959
  & 111640.48  & 11.623
  & 855.97     & 6.753
  & 140097.29  & 11.850 \\

3 & 19192.97   & 9.862
  & 185658.64  & 12.132
  & 2370.11    & 7.771
  & 228871.67  & 12.341 \\

4 & 40711.96   & 10.614
  & 276446.47  & 12.530
  & 5495.96    & 8.612
  & 349862.14  & 12.765 \\

5 & 78050.96   & 11.265
  & 395701.12  & 12.888
  & 9941.82    & 9.205
  & 522521.09  & 13.166 \\

6 & 143170.45  & 11.872
  & 568338.58  & 13.250
  & 18343.81   & 9.817
  & 769205.62  & 13.553 \\

7 & 258134.44  & 12.461
  & 859906.26  & 13.665
  & 35996.98   & 10.491
  & 1155369.80 & 13.960 \\

8 & 484755.09  & 13.091
  & 1465415.44 & 14.198
  & 115656.06  & 11.658
  & 1793458.22 & 14.400 \\

9 & 1592662.96 & 14.281
  & 5639944.02 & 15.545
  & 509093.26  & 13.140
  & 5163125.70 & 15.457 \\

\hline
\end{tabular}
\end{table}


\end{document}